# Apple Vision Pro for Healthcare: "The Ultimate Display"?
## *Entering the Wonderland of Precision Medicine*

Jan Egger[1,2,3*], Christina Gsaxner[1,4,5], Xiaojun Chen[6,7], Jiang Bian[8], Jens Kleesiek[1,3,9,10], Behrus Puladi[4,11*]

[1]Institute for Artificial Intelligence in Medicine (IKIM), Essen University Hospital (AöR), Girardetstraße 2, 45131 Essen, Germany
[2]Center for Virtual and Extended Reality in Medicine (ZvRM), Essen University Hospital (AöR), Hufelandstraße 55, 45147 Essen, Germany
[3]Cancer Research Center Cologne Essen (CCCE), University Medicine Essen (AöR), Hufelandstraße 55, 45147 Essen, Germany
[4]Department of Oral and Maxillofacial Surgery, University Hospital RWTH Aachen, Pauwelsstraße 30, 52074 Aachen, Germany
[5]Institute of Computer Graphics and Vision, Graz University of Technology, Inffeldgasse 16/II, 8010 Graz, Austria
[6]Institute of Biomedical Manufacturing and Life Quality Engineering, State Key Laboratory of Mechanical System and Vibration, School of Mechanical Engineering, Shanghai Jiao Tong University, Dongchuan Road 800, Minhang District, Shanghai 200240, China
[7]Institute of Medical Robotics, Shanghai Jiao Tong University, Dongchuan Road 800, Minhang District, Shanghai 200240, Shanghai, China
[8]Health Outcomes & Biomedical Informatics, College of Medicine, University of Florida, Gainesville Florida 32610, United States
[9]German Cancer Consortium (DKTK), Partner Site Essen, Hufelandstraße 55, 45147 Essen, Germany
[10]Department of Physics, TU Dortmund University, August-Schmidt-Str. 4, 44227 Dortmund, Germany
[11]Institute of Medical Informatics, University Hospital RWTH Aachen, Pauwelsstraße 30, 52074 Aachen, Germany
*Corresponding authors: jan.egger@uk-essen.de (J.E.) and bpuladi@ukaachen.de (B.P.)

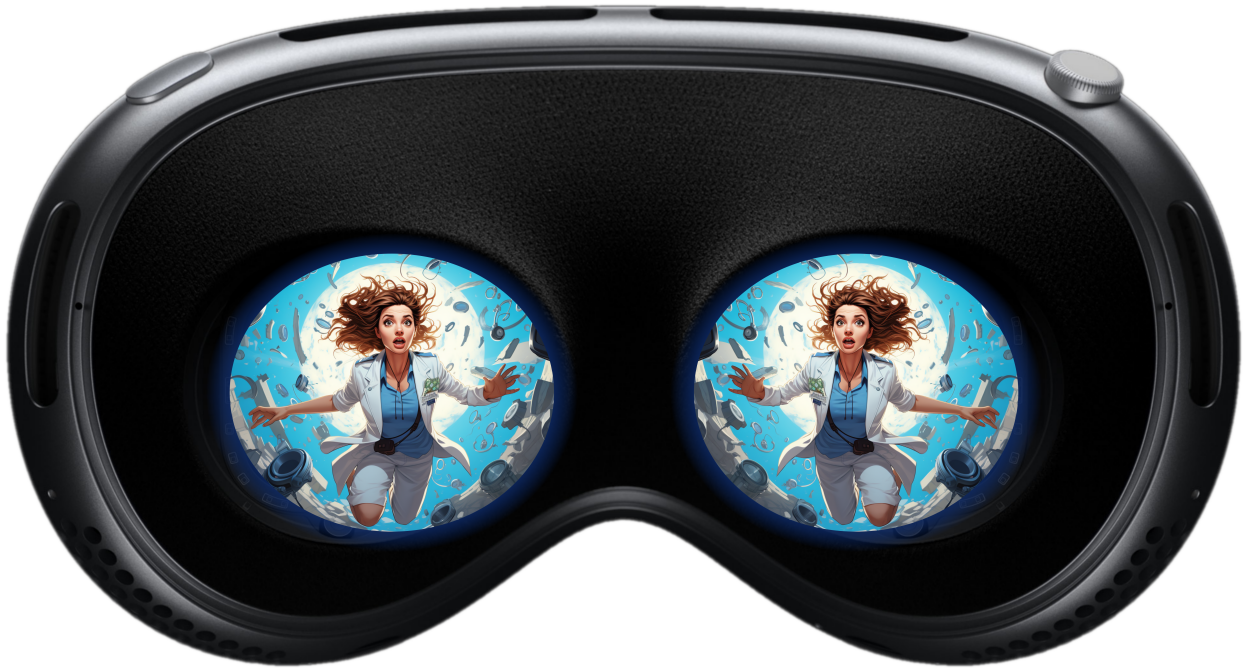

**Abstract**: At the Worldwide Developers Conference (WWDC) in June 2023, Apple introduced the Vision Pro. The Vision Pro is a Mixed Reality (MR) headset, more specifically it is a Virtual Reality (VR) device with an additional Video See-Through (VST) capability. The VST capability turns the Vision Pro also into an Augmented Reality (AR) device. The AR feature is enabled by streaming the real world via cameras to the (VR) screens in front of the user's eyes. This is of course not unique and similar to other devices, like the Varjo XR-3. Nevertheless, the Vision Pro has some interesting features, like an inside-out screen that can show the headset wearers' eyes to "outsiders" or a button on the top, called "*Digital Crown"*, that allows you to seamlessly blend digital content with your physical space by turning it. In addition, it is untethered, except for the cable to the battery, which makes the headset more agile, compared to the Varjo XR-3. This could actually come closer to the "*Ultimate Display*", which Ivan Sutherland had already sketched in 1965. Not available to the public yet, like the Ultimate Display, we want to take a look into the crystal ball in this perspective to see if it can overcome some clinical challenges that - especially - AR still faces in the medical domain, but also go beyond and discuss if the Vision Pro could support clinicians in essential tasks to spend more time with their patients.

**Introduction**

Ivan Sutherland, the pioneer of Computer Graphics and developer of the first Augmented Reality (AR) Head-Mounted Display (HMD) as early as the early 1960s, and before the first Personal Computer (PC), outlined what Apple's Vision Pro could come close to in his much-cited 1965 paper "*The Ultimate Display*" [1]. The Vision Pro from Apple is able to cover the wide range of the Reality-Virtuality (RV) continuum as explained by Milgram et al. [2] (Figure 1): From the *reality* (left), by streaming only the real world without additional information to the user, over a *Mixed Reality* (MR, middle) by adding digital information to the real world known as AR and vice versa (i.e., Augmented Virtuality), to a complete immersion aka Virtual Reality (VR, right), by showing the user only generated virtual content. For VR, there are currently more and lower priced headsets, e.g., from Meta, Vive, Varjo, PlayStation, Google (the cheapest is probably the Google Cardboard, which has been discontinued, or a variant from another vendor) available than for AR. In general, AR faces the challenge of also analyzing the *reality* to augment it at the *right* position with digital content for the user. An exception is AR used for a *pure* simulation, where the AR hologram is indeed shown in the real world, but has no meaningful relation or interactions with real-world objects. An example of such a scenario is a surgical simulation [3]. The AR hologram is shown in front of the user to inspect and interact with it. Still, the underlying hardware and software need to capture and model the real world to *anchor* the hologram within it, so that the hologram does not *drift* away from the user's view. Either way, AR can enable a physician x-ray vision into a patient [4] in the medical domain, e.g., during a surgical intervention by showing the tumor as hologram with a spatial perception inside the patient, before even opening the patient. In general, patient information, like preoperative imaging, is displayed on PC monitors, which leads to the so-called "*switching focus problem,*" where a physician has to divide his or her attention between the patient and the digital information (of the patient). Hansen et al. [5] outlined that such a division leads to an increased mental workload, disorientation, and deteriorated hand-eye coordination. AR can tackle these problems by overlaying the digital (patient) information directly onto the patient, like a hologram. A recent systematic review of the HoloLens (Microsoft's AR headset) in medicine [6], showed the broad applications it can be applied to and the massive research efforts during the last years in all these areas (which should ultimately lead to the development of an intelligent healthcare metaverse [7], [8]). Hence, we want to highlight the user groups and applications that have been approached with AR in the next sections and discuss and hypothesize if the remaining challenges can be overcome with the (new) Vision Pro capabilities from Apple.

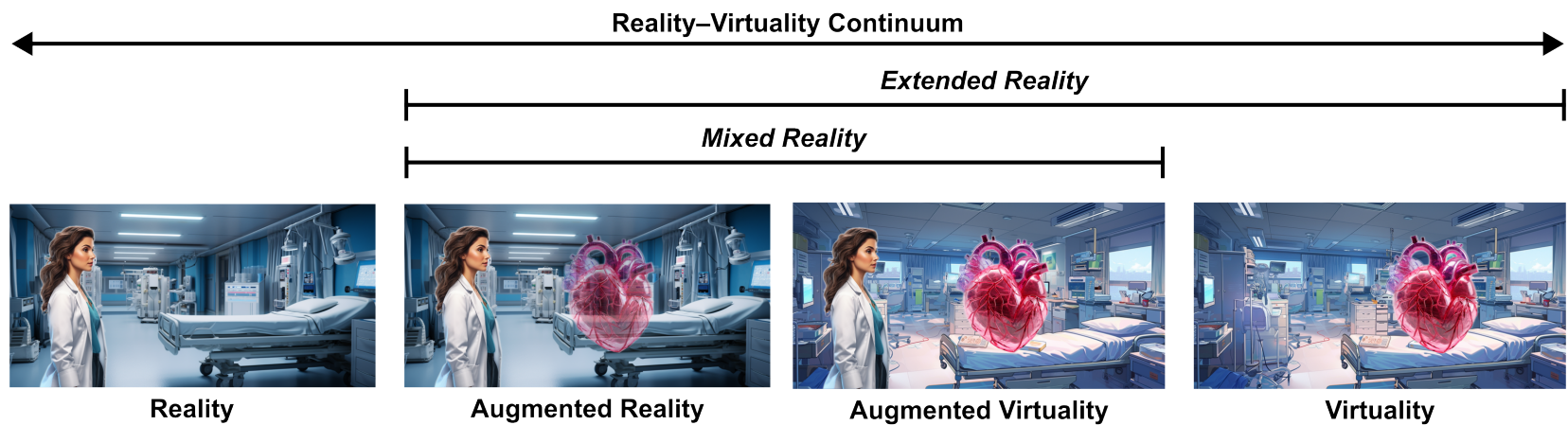

**Figure 1.** The Reality-Virtuality (RV) continuum as explained by Milgram et al. [2]. From left to right: An (real) operation room (Reality), hologram of a patient heart in a (real) operation room (Augmented Reality), a (real) physician inside a computer-generated operation room (Augmented Virtuality) and a computer-generated operation room with no real objects or persons (Virtuality aka Virtual Reality).

**Medical AR user groups and their challenges**

AR applications for physicians and healthcare professionals cover by far the largest user group [6]. However, especially for AR-supported image guidance and navigation, very high accuracy and reliability may be needed [9]. Applications, for which sub-millimeter precision is not necessary, are, for example, ablations, ventriculostomy [10-14] or certain orthopedic interventions [15]. Here, the HoloLens is with its inside-out tracking already a promising tool, but for applications that need, for example, sub-millimeter precision, it cannot be used reliably yet. An example is the deep brain stimulation (DBS) procedure used for treating essential tremor and Parkinson's disease, where millimeter-to-submillimeter accuracy in DBS targeting (an electrode placement inside the brain) can be important [16]. Another study exploring the clinical accuracy of the HoloLens for neuronavigation concludes also that it is currently not within clinically acceptable levels [17]. The same holds true for some application scenarios in orthopedic surgery [18], like screw placement, where there is still room for improvement [19]. We expect the Vision Pro to move the needle in terms of accuracy, because of its strong inside-out tracking through 12 built-in cameras and LiDAR (Light Detection and Ranging) sensing method, which is the key aspect for increasing the accuracy of AR. We do not see the often-criticized end-user price of $3,499 (without tax) for the Vision Pro as an issue for this user group. The price is similar to the HoloLens and much cheaper at a fraction of the costs compared to existing, and clinically used conventional medical navigation systems, e.g., from Brainlab or Medtronic. In this context, it is important to mention that the use of AR head-mounted displays (HMDs) with direct patient reference (and outside of research in the context of feasibility studies) requires an appropriate regulatory approval and a certification/classification as medical device. The Magic Leap 2, for example, received certification for usage in an operating room in January of this year (2023). Moreover, and especially in a medical context, the display may also require an image focus in surgical table distance [20].

Students are the second most common intended user group [6] with applications, for example, in educational training, like the *HoloPointer*, a virtual AR pointer for laparoscopic surgery training [21]. Another example is the usage of MR to teach medical students catheter placement [22] or a phantom experiment to study the effectiveness of learning using AR in the access of the central venous [23]. We found that the effects of HoloLens-based learning compared to conventional learning, e.g., by using cadavers or other computerized methods, seem to be rather small. A reason for this is that the learning tools usually used are relatively simple, conventional 3D models and more innovative visualizations, including interactive, dynamic content, which cannot be easily delivered by regular computerized methods, have not been explored in depth yet [6]. Here, we believe that the Vision Pro can raise the bar, because of its high-resolution visualization (consisting of two micro-OLED displays with a total of 23 megapixels) that uses eye-tracking to make the headset also usable as a desktop screen with its fine textual details. In addition, the 3D user interface, using precise and intuitive finger tracking, can be a game changer. Despite these advances, it remains to be seen if an AR headset has a major advantage over a conventional screen. The cost of the Vision Pro is also a concern, especially for students with a limited financial budget, but the "Pro" might indicate that there will also be a "regular" version released.

Patients are currently the least frequented target user group with AR devices [6]. An example is *MemHolo*, which provides MR experiences for subjects with Alzheimer's disease [24]. Other examples in this area include the usage of the HoloLens as an assistant and monitoring tool for medication adherence [25] and the usage of a HoloLens-based system for functional mobility assessment [26]. However, many interesting assistance and monitoring applications are limited by the restricted possible usage time of untethered AR headsets primarily due to battery life, and this problem seems to remain with the Vision Pro, which has a similar battery life of approximately two hours, like the HoloLens. The only consolation is the "*external*" battery pack that can be swapped for the Vision Pro. Finally, the Apple M2 and the new R1 chip of the Vision Pro could potentially boost the relatively small number of applications that have been explored so far, e.g., by reducing motion sickness, especially for elderly patients, which are more susceptible to MR sickness than younger users.

In summary, we identified the main bottleneck for establishing AR in clinical practice to be the hardware, its limitations and the special requirements of medical regulations. This is hard to overcome for (software) developers. An option is to combine an AR device with external tracking systems to increase the preciseness [27]. On the downside, such combinations require complicated setups and the calibration of bulky (external) devices, which takes away the "*lightness*" of purely head-worn systems. However, we believe that the Vision Pro can provide significant advancements as the new state-of-the-art device, with much-needed improvements in precision, reliability, usability, workflow, and perception. The combination of AR/VR has the potential to transform healthcare by aiding in diagnostics, improving surgical procedures, facilitating remote patient care, and enhancing medical training. Many other novel applications may emerge because of the advancements in its hardware and software platform.

**Vision Pro features**

Apple announced some interesting (and still unique) features for the Vision Pro. In the upcoming sections, we want to introduce and discuss these in the context of medical MR with a focus on AR:

*Digital Crown* – The Digital Crown is a turning button that allows you to easily and seamlessly blend digital content with the physical environment surrounding you. It definitely provides an easy option to quickly *escape* VR without the need to remove the headset. Depending on the capabilities and accessibility for developers to the Digital Crown, it could also enable Diminishing Reality (DR) [28] for users to remove distracting real-world objects or environments for physicians, students, or patients, to reduce visual overstimulation (it could even diminish/remove the headsets of other users). Note that DR needs, in general, a Video See-Through (VST) capability so far and cannot be realized by Optical See-Through (OST), like the HoloLens. This is similar to the medical HMD from VOSTARS, who have combined an OST, i.e., the view through a semi-transparent glass with a VST function [29]. VOSTARS has realized this by allowing the semi-transparent glass to set itself into a non-transparent mode. In this context, the Vision Pro distinguishes itself, because it is not an OST versus VST, but a VST versus VR HMD, i.e., walking on the VR continuum, which Paul Milgram and Fumio Kishino postulated in 1994 [30].

*Virtual eyes* – Spooky at first sight, the outward display reveals a user's eyes while wearing the Vision Pro. According to Apple, this feature is meant to let others know when the Vision Pro wearer is using apps or is fully immersed in a virtual world. This feature is not needed in classical AR devices, like the HoloLens, because the headset wearer's eyes can be seen through the transparent display. On the contrary, the "*virtual eyes*" feature can enable DR in healthcare, with human-like characteristics for the patients. In the end, we are talking about very seriously ill humans, whom we do not want to put an additional strain on, by being treated by someone hidden behind a screen.

*3D camera* – The Vision Pro also features a 3D camera, which allows a user to capture spatial photos and spatial videos in 3D. Being more of a gimmick for non-professional end users, this could be quite useful for the professional (healthcare) sector. We are thinking of a 3D documentation of interventions or documenting at least certain crucial steps of an intervention. This is certainly also interesting for training new residents that can reenact treatments in 3D, even from the view of the wearer. It could also serve as an investigation of malpractice. The 3D capabilities can also provide an alternative or an addition to 3D documentation of crime and crash scenes, which is common practice during forensic and medicolegal investigations nowadays [31].

*Interaction* – According to Apple, a user can interact with virtual content by simply looking at an element, tapping the fingers together to select, and using the virtual keyboard or dictation to type. There also seems to be a "*Visual Search*" function, which may be similar to the "*Visual Look Up*" feature found on iPhones and iPads. "*Visual Search*" will allow users to interact with items and text, for opening webpages and translation into other languages in real-time. Interestingly, this was already described by Ivan Sutherland as "*language of glances*" to interact with a computer, with an example where we look at a corner of a triangle that becomes then rounded [1]. Looking something up, is definitely interesting, e.g., for surgeons who have their hands full with (surgical) instruments, or nonverbal users.

*Optic ID* – The Vision Pro is supposed to support an optical identification of a user for authentication, like the iPhone's Face ID. The biometric method is enabled by scanning the iris of the user wearing the headset. This is not unique to the Vision Pro, because the HoloLens 2 already supported an iris authentication. However, this is also an important feature for the Vision Pro in healthcare, where you have sensitive patient data and unlocking the device and giving access to the data should only be permitted to authorized users. Furthermore, it enables a hands-free login in a possible sterile environment.

*ZEISS optical inserts* – The Carl Zeiss AG is working together with Apple to provide precision optics for users who require vision correction. This will enable a better visual experience, better comfort (because a user does not need to wear his or her prescription glasses and frames inside the Vision Pro with its limited space) and may also avoid side effects like "overheating", headaches, etc. The additional lenses can be attached magnetically to the main lens, which makes an insertion and replacement easy.

*Head strap* – The head strap of the Vision Pro seems to be easily exchangeable and there are already several alternatives (materials, colors, etc.) available (online). This will ultimately enable more comfort, in comparison to the HoloLens and the Varjo XR-3, which are adjustable, but still consist of quite stiff (plastic) frames around the head. It will definitely be more comfortable like the "*Sword of Damocles*" by Ivan Sutherland and his student Bob Sproull in 1968, where the user had to be strapped into the AR/VR HMD device, which was suspended from the ceiling, because it was so heavy. The head strap contains also the Vision Pro's speaker, which is placed directly over the user's ears and is supposed to virtualize surround sound. However, the strap means that all main components are integrated with the visor. The HoloLens 2, for example, distributes the weight more evenly, by integrating components in a plastic case located at the rear, including the battery and system on a chip (SoC) board. This was done by Microsoft based on their experiences with the HoloLens 1, to make the HoloLens 2 more comfortable. On the contrary, the Vision Pro seems to be worn more like a VR device that fits tightly onto the face, instead of partly hovering in front of the user's eyes, such as other AR devices which are more similar to prescription glasses. The Vision Pro will be in a similar weight range (~500 g) to other common AR/VR headsets, despite not integrating the battery into the device and separating it. In part, this could be attributed to the material, e.g., using an aluminum alloy frame.

*3D persona avatars* – During the setup of the Vision Pro, the user's face is scanned by the headset to create a photorealistic avatar that is then used by the device's operating system, called visionOS. For better fitting purposes, the user's face can also be scanned with the TrueDepth camera of an iPhone as seen at a public demo at the Worldwide Developers Conference (WWDC) in June 2023. Furthermore, the user's ears can be scanned to optimize the speakers that are inside of the head strap, because, aside of the shape and size of a head, the ear sizes, positions, and distances can vary a lot between users. All these user-specific options will make the headset more comfortable to wear and this will finally increase the acceptability of the new device, not only in the healthcare sector.

*Unity* – The Unity engine already supports pretty much all common headset devices, may it be AR, VR or MR, like the Oculus, HoloLens, Google VR and also the OpenVR API (Application Programming Interface) that has already been used in medicine [32]. Thus, the support of Unity of the Vision Pro will make the development and porting of apps much easier. This also makes a comparison and evaluation between devices with the "same" app easier. Overall, Unity support will enable a faster development of apps, hopefully fast enough to reach a critical mass, before end users may turn their backs, which is needed for the long-term success of the new Apple device.

*Medical device* – Talking about the healthcare sector, we need to mention that the Vision Pro is not a medical device (yet). We are also not aware that Apple is planning in this direction, at least not publicly. Apple, however, promoted *Complete HeartX* as an education app for medical students, by providing hyper-realistic 3D models and animations of the heart and medical issues. In comparison, Microsoft advertised the HoloLens 2 specifically for medical scenarios, e.g., with a plenary presentation of Bernard Kress (that time being Principal Optical Architect, HoloLens team, Microsoft Corp.) at the SPIE Medical Imaging in February 2020. Nevertheless, we are sure that the Vision Pro will be used by researchers and companies for medical applications, like other devices, if Apple does not explicitly prohibit it somehow, e.g., by monitoring their devices and locking them. Interesting will be the integration of the Vision Pro with other software and hardware commonly used in clinical settings, e.g., an integration with Electronic Health Records (EHRs) [33, 34].

**Conclusion and final remarks**

Summarized, we see Apple entering the MR headset market with its own device very positively. Another company making a commitment can only benefit the technical progress in this area, not only from a hardware side but also from the software perspective, by enabling new user input concepts and apps. With the current hardware specifications available, and Apple having a history of delivering cutting-edge devices and beyond, there is a very good chance of another AR "evolution" or even "revolution" (Table 1). This is a very important aspect, because (1) AR devices are sparse in comparison to VR, and (2) the product cycle to a possible successor is in the range of years, which feels like ages when compared to the current pace of technological developments, especially in (medical) Computer Science [35]. An example is the first version of the HoloLens, which was introduced by Microsoft in 2016. It took Microsoft over three years for its successor, the HoloLens 2 (released in limited numbers on November 7, 2019).

It was in 2007 when Steve Jobs released the iPhone, creating a new category of smartphones. What made it special was that Apple combined existing technologies and software in a way that made previously unexpected things possible, which Apple repeated with the iPad in 2010 and then with the Apple Watch in 2015. Hence, the Vision Pro could lead to the breakthrough of AR/VR/MR, but also DR (because it's a passthrough technique), due to its enormous technological progress. Moreover, the technological progress can not only enable more precise, patient-individual treatments, but also making them more efficient, e.g., by hands-free, instant authentications, documentations and search-queries and their results shown directly within the field of vision of the HMD user. Further, well-engineered AR can literally help to overcome communication barriers in a face-to-face manner. Thus, supporting clinicians in a range of essential tasks and reducing the administrative burden on clinicians to allow them to spend more time with their patients, which is also a key challenge for upcoming Artificial intelligence (AI)-based foundation models [36].

The recent rise of Large Language Models (LLMs) [37], like ChatGPT, can also not be ignored in the context of the upcoming Vision Pro, because it changed how we communicate with computers and the Vision Pro can certainly be enhanced with an integration of a ChatGPT-like bot. Even still in its infancy, when it comes to hard facts, especially in healthcare [38, 39], ChatGPT enables already a fluent, partly facts-based, conversation. This will, in the short or long run, also fully enter the headset market, where someone can discuss upcoming treatment steps on a human-like level based on a trained LLM. However, we do not think that this will be specific to the Vision Pro and with ChatGPT, Microsoft seems to currently have the edge.

Nevertheless, digitization has promised to make medical care more efficient. But will screens come between doctors and patients [40]? At the same time, for every hour doctors spend with their patients, they spend two hours at the computer [41]. Whether the Apple Vision Pro will blur the line between patient encounters and computer time also depends on patient and physician acceptance. In ophthalmology, the Vision Pro is seen as a breathtaking application [42]. However, this must first be confirmed in clinical trials with rigorous scientific methodology and hard clinical endpoints. When the Vision Pro is used for patient scenarios, the focus must be on the patient, not the technology, which is always a vehicle for existing problems.

To sum up, the Vision Pro enriches the current headset market and provides another alternative, especially in the limited AR segment. The Vision Pro will be likely at the technological forefront in this sector, which puts pressure on other vendors to draw level and surpass to keep competitive. The price is for entertainment consumers certainly a burden, but for professionals, especially in the healthcare sector, with the Microsoft HoloLens being in the same price range and the Varjo XR-3 even double that amount, not a deal-breaker. And this gets even more *irrelevant* if seen in relation to the other hardware costs in healthcare, such as navigation systems, imaging equipment (Computed Tomography (CT) and Magnetic Resonance Imaging (MRI) scanners), which can cost millions. In Ivan Sutherland's sense, it is not yet the *Wonderland* that Alice or the Doctors (added by us) walked into, but perhaps a bit closer.

**Acknowledgements**

This work was supported by the REACT-EU project *KITE* (*Plattform für KI-Translation Essen*, https://kite.ikim.nrw/, EFRE-0801977), FWF *enFaced 2.0* (KLI 1044, https://enfaced2.ikim.nrw/), NIH/NIAID R01AI172875 and NIH/NCATS UL1 TR001427. Behrus Puladi was funded by the Medical Faculty of the RWTH Aachen University in Germany as part of the Clinician Scientist Program. Christina Gsaxner was funded by the Advanced Research Opportunities Program (AROP) from the RWTH Aachen University. Furthermore, we acknowledge the *Center for Virtual and Extended Reality in Medicine* (*ZvRM*, https://zvrm.ume.de/) of the University Hospital in Essen, Germany. Finally, we want to thank ZhaoDi Deng, Ahmad Idrissi-Yaghir, Rainer Röhrig and Midjourney.

**Table 1.** Side-by-side comparison of the specifications of several existing Mixed Reality (MR) headsets with the upcoming Apple Vision Pro.

| Parameter | Apple Vision Pro | Pico 4 | Meta Quest 3 | HTC Vive XR Elite | Varjo XR-3 | HoloLens 2 | Magic Leap 2 |
|---|---|---|---|---|---|---|---|
| AR | VST | VST | VST | VST | VST | OST | OST |
| VR | yes | yes | yes | yes | yes | no | no |
| Resolution (per eye) | 3800 x 3000 | 2160 x 2160 | 2064 × 2208 | 1920 × 1920 | 1920 x 1920 focus, 2880 x 2720 peripheral | 2048 × 1080 | 1440 x 1760 |
| Refresh Rate | 90 Hz | 90 Hz | 120 Hz | 90 Hz | 90 Hz | 120 Hz | 60 Hz |
| Eye, head & hand tracking | yes | yes | yes | yes | yes | yes | yes |
| Depth sensing | LIDAR + TrueDepth | Depth camera | Depth camera | Depth camera | LiDAR + RGB fusion | ToF Depth | iToF Depth |
| RAM | up to 16 GB | 8 GB | 12 GB | 12 GB | | 4 GB | 16 GB |
| Storage | 64 GB | 128 / 256 GB | 128 / 256 / 512 GB | 128 GB | | 64 GB | 256 GB |
| Battery life | 2 h | 2 h | 3 h | - | - | 2 h | 3.5 h |
| Weight | ~500 g (estimated, without external battery) | 295 g + headband 291 g | 509 g | 625 g | 594 g + headband 386 g | 566 g | 260 g |
| Untethered | Yes | Yes | Yes | No | No | Yes | Yes |
| Price | $3,499 | $430 | $499 | $1,100 | $6,500 | $3,500 | $3,299 |

VST, Video See-Through; OST, Optical See-Through.